\documentclass[
]{ceurart}

\usepackage{listings}
\usepackage{fvextra}
\usepackage{tcolorbox}
\DefineVerbatimEnvironment{Verbatim}{Verbatim}{breaklines=true}

\usepackage{soul}
\usepackage{tikz}
\definecolor{color1}{HTML}{93003a}
\definecolor{color2}{HTML}{cf3759}
\definecolor{color3}{HTML}{f4777f}
\definecolor{color4}{HTML}{ffbcaf}
\definecolor{color5}{HTML}{ffffe0}
\definecolor{color6}{HTML}{a5d5d8}
\definecolor{color7}{HTML}{73a2c6}
\definecolor{color8}{HTML}{4771b2}
\definecolor{color9}{HTML}{00429d}
\newcommand*{\opacity}{33}
\newcommand*{\minval}{0.000}
\newcommand*{\maxval}{1.0}

\newcommand{\gradient}[1]{
    \ifdimcomp{#1pt}{>}{\maxval pt}{#1}{
        \ifdimcomp{#1pt}{<}{\minval pt}{#1}{
            \pgfmathparse{int(round(8*(#1/(\maxval-\minval))-(\minval*(8/(\maxval-\minval)))))}
            \xdef\tempa{\pgfmathresult}
            \ifcase\tempa
                \cellcolor{color1!\opacity} #1\or
                \cellcolor{color2!\opacity} #1\or
                \cellcolor{color3!\opacity} #1\or
                \cellcolor{color4!\opacity} #1\or
                \cellcolor{color5!\opacity} #1\or
                \cellcolor{color6!\opacity} #1\or
                \cellcolor{color7!\opacity} #1\or
                \cellcolor{color8!\opacity} #1\or
                \cellcolor{color9!\opacity} #1
            \fi
    }}
}

\begin{document}

  \copyrightyear{2024}
\copyrightclause{Copyright for this paper by its authors.
  Use permitted under Creative Commons License Attribution 4.0
  International (CC BY 4.0).}

\conference{CHR 2024: Computational Humanities Research Conference, December 4–6, 2024, Aarhus, Denmark}

\title{On Classification with Large Language Models\\in Cultural Analytics}

 \author{David Bamman}[%
 orcid=0009-0003-1171-9408,
 email=dbamman@berkeley.edu,
 ]
 \address{School of Information, UC Berkeley}

 \author{Kent K. Chang}[%
 orcid=0009-0008-6430-3701,
 email=kentkchang@berkeley.edu,
 ]

 \author{Li Lucy}[%
 orcid=0000-0002-6021-7370,
 email=lucy3_li@berkeley.edu,
 ]

 \author{Naitian Zhou}[%
 orcid=0009-0005-1991-2258,
 email=naitian@berkeley.edu,
 ]

\begin{abstract}
  In this work, we survey the way in which classification is used as a sensemaking practice in cultural analytics, and assess where large language models can fit into this landscape.  We identify ten tasks supported by publicly available datasets on which we empirically assess the performance of LLMs compared to traditional supervised methods, and explore the ways in which LLMs can be employed for sensemaking goals beyond mere accuracy. We find that prompt-based LLMs are competitive with traditional supervised models for established tasks, but perform less well on \textit{de novo} tasks. In addition, LLMs can assist sensemaking by acting as an intermediary input to formal theory testing.
\end{abstract}


\maketitle

\section{Introduction}

One of the core tasks in cultural analytics is the act of \emph{classification}.  We sort discrete things---books, passages, sentences, works of art, songs---into buckets whose boundaries have either been established by long theoretical traditions (such as categories of poetic form\ \citep{long2016literary,walsh2024sonnet}), or that are newly circumscribed in the course of our research (cf. ``technological strangeness''\ \citep{simeone2019towards}).  We carry out this act for several reasons: to test the association of features with those categories in order to better understand the categories themselves; to recreate human judgment at a scale beyond what people are able to carry out alone; and to problematize original category boundaries, leveraging machines to help us poke holes in our current understanding of them. As models, classifiers ``return us to the process \ldots through which we construct our knowledge of phenomena that exceed our direct observation'' (\citet[651]{piper2017think}).

Each one of these goals requires different machinery to accomplish it, from  interpretable linear models that allow us to isolate the effect of an attribute on a category, to complex models that may sacrifice such interpretability for predictive accuracy in those cases when accurate judgments are all that matter.

As large language models (LLMs) have arisen to become the state of the art for a wide range of tasks in natural language processing, a growing body of work is investigating the tradeoffs they bring for this act of classification in fields for which text is data\ \citep{walsh2024sonnet,ziems2024can,rebora2023comparing}. High-quality LLMs are generally far more expensive than alternatives (in compute, environmental, and financial costs), and come burdened with a Pandora's box of pre-training---their source of state-of-the-art performance and general-purpose knowledge, but also the vector along which undesirable knowledge is carried. At the same time, they offer hope for dramatically reducing the amount of human effort required for meaningful humanistic inquiry\ \citep{ziems2024can}---trading the scale of manual annotation often required by supervised models for more focused injection of human and domain knowledge into prompts.

To shed some light on the contours of this landscape, we investigate the intersection of LLMs, classification and cultural analytics along four dimensions:

\begin{itemize}
    \item First, we survey recent work published in cultural analytics (CHR, \emph{Journal of Cultural Analytics}, NLP conferences, etc.) to identify the uses to which researchers are currently employing classification. We group those uses into a typology of six distinct categories.
    
    \item Second, we identify a set of ten tasks supported by publicly available datasets on which we can examine tradeoffs for some of those uses.  These tasks represent a wide range of phenomena of interest to cultural inquiry and provide a sample of the real ways in which researchers now are using classification in their work.  We focus on English-language datasets only in this work; our results are therefore limited to the English-language affordances of downstream applications.
    
    \item Third, we use that suite of datasets to benchmark the performance of different classification models, including bag-of-words linear models, masked language models (such as BERT and RoBERTa) and large language models (both prompting and fine-tuning).  We find prompt-based LLMs to be competitive with traditional supervised models for established English-language tasks, but lag behind supervised models for \emph{de novo} tasks. While cautioning against the accuracy-maximizing incentives that benchmarks create\ \citep{orr2024ai}, this suite of datasets allows us to assess the tradeoffs between computation, cost and accuracy for several of those use cases where classification accuracy is paramount.

    \item Fourth, we carry out an exploratory case study bringing together LLMs with this suite of datasets for a goal that does not seek simply to maximize classification accuracy: \emph{category sensemaking}.  While this study can only be suggestive, we find that LLMs provide a mechanism for exploratory data analysis that complements existing methods.

\end{itemize}

We release data for others to explore these tradeoffs themselves; data and code to support this work can be found at \url{https://github.com/bamman-group/ca-classification-data}.

\section{How are we using classification?}

As distinct from mere automation, the use of classification in cultural analytics is often employed for the purpose of sensemaking.
At a high level, the act of supervised classification entails learning a function $\hat h(x)$ from an input source $x$ (e.g., text) to a choice (or choices) $y$ drawn from an enumerable output space $\mathcal{Y}$ (e.g., for genre classification, $\mathcal{Y} = \{\textrm{science fiction, western, detective story}, \ldots\}$), though for our purposes, we also include regression problems (where $y \in \mathbb{R}$). Classification requires the existence of a previously selected set of categories $\mathcal{Y}$, which is a political choice grounded in a system of knowledge\ \cite{foucault94}; as Bowker and Star note, ``each category valorizes some point of view and silences another''\ \citep[5]{bowker}.  This is an important subject of critique not only when the objects of study are people\ \cite{d20204,benthall2019racial} but also in recognizing that classification can only make statements with respect to the categories that are classified (and not to the gaps that exist between them).

For much work in this space, the representation of $x$ is critical---e.g., the choice of representing $x$ as a vector corresponding to word counts from a defined vocabulary, or other more specialized features (e.g., type-token ratio).  To be precise, we can describe this representation as a function of its own (e.g., $f(x)$), which is the direct input to a classification function ($\hat h(f(x))$), though modern neural classifiers convolve both functions into one learned operation.

We can separate out these uses of classification into distinct yet inter-related categories:

\paragraph{1. Category sensemaking.}

One of the most widely used applications of classification in cultural analytics is understanding the characteristics of a category through the features that are predictive of it.  We can view this as a sensemaking exercise, whose main degree of freedom is the choice of representation function $f(x)$. This function can vary from a kitchen-sink approach (throw every feature into a model and learn what is important) to those that define specific, theory-driven feature classes.  We see many examples of this kind of sensemaking.  \citet{long2016turbulent} use a featurized model to identify the defining characteristics of stream-of-consciousness form---above all, type-token ratio---and how those defining features both converge and diverge across English and Japanese stream-of-consciousness texts.   \citet{simeone2019towards} use classification to uncover the characteristics of technological \emph{nova}---moments in science fiction texts where new technologies are introduced; using a bag-of-words featurization, they find that strangeness is characterized not merely by diction (the choice of words, such as ``ship'') but also by morphosyntax (presence of determiners and prepositions).  Several studies use the machinery of classification to interrogate the boundaries of gender performance in English\ \citep{jockers2016understanding,Underwood2018Transformation,cheng2020fleshing} and French\ \citep{vianne2023}. 
\citet{bode2018} uses classification in combination with topic modeling to learn that the theme of ``nonmetropolitan colonial spaces'' most strongly differentiates Australian fiction from British and American fiction.

\paragraph{2. Top-down theory testing.}

A second, but related, use of classification is explicitly theory-driven: articulating a formal test (often of association between a feature and a category) and carrying out that test.   We can differentiate this use from exploratory sensemaking by the falsifiability of its claims: a theoretical claim is articulated,   and a corresponding null hypothesis is either rejected or not when evaluated on data. For \citet{piper2022toward}, theory provides the boundaries of operationalization (in defining the feature space through which a text is represented), and classification provides a test for whether that operationalization is sufficiently descriptive to differentiate texts that are narrative from those that are not across a variety of contexts.  The boundaries between this use case and that of category sensemaking are often fluid---even within this theory-driven choice of feature space, \citet{piper2022toward} also engage in sensemaking by identifying a \emph{minimal} set of features from predictive models that characterize narrativity and ground that minimal set in a more precise ``distant worlds'' theory of narrative.  \citet{steg-etal-2022-computational} take a similar theory-driven approach, testing the degree to which measures of reader suspense, surprise and curiosity are predictive of narrativity.

\paragraph{3. Challenging category boundaries.}

A third use, related to use \#1, employs classification to challenge the initial category boundaries of a concept.   
\citet{long2016literary} is an exemplary use of this; this work uses classification primarily to explore and challenge the category boundaries of haiku by examining misclassifications---cases where a model classifies a non-haiku poem as haiku and vice versa.  They additionally engage in close reading of the model itself, by identifying salient features of texts that are statistically associated with haiku and find that the language of natural imagery and brevity are strong indicators of the form (beyond simple syllabic patterns).
Work by \citet{broadwell2017tell} likewise trains a folktale type classifier on Danish narratives to identify ``liminal stories that shift across existing categories'' (23); these stories expose the ambiguity of the initial categories and highlight the ways in which their boundaries are permeable.  

\paragraph{4. Category coherence.}

While use \#3 assumes a fixed category object whose boundaries are unknown, a fourth use employs classification accuracy as a proxy for the coherence of a concept itself, under the assumption that statistical regularities could only exist for concepts that are somehow real. Much work in the first three uses carries with it an implicit theoretical test in the coherence of the category; \citet[350]{long2016turbulent} argue that the predictability of stream-of-consciousness texts vs. realist texts provides evidence for ``some of degree of formal unification''; 
\citet{thompson2017computational} use classifier accuracy at recognizing images from the Dada art movement as evidence of its coherence as a concept (``there is substance behind the name Dada,'' 193) while stressing that such coherence is not absolute: classifier mistakes illustrate ``the porous boundaries of the category'' (194).
In a similar vein, the counterfactual provides information about the dissolution of the construct. \citet{Underwood2018Transformation} predict character gender from the attributes and actions associated with characters, and find that decreasing predictive accuracy over time suggests an increasing ``instability'' in the gender construct.

\paragraph{5. Search.}

We might term a fifth category, search, as ``classification-assisted close reading.''  This line of work uses the machinery of a classifier as a finding aid to zero in on passages that instantiate a target category.  \citet{wilson2023living} is an exemplary use of this approach.  This work explores the concept of ``atypical animacy'' (machines that are depicted as animate vs. non-animate) in a large  collection of historical books, newspapers and journal articles. After designing a method to identify passages of machine animacy in order to make them accessible to close reading, this work uses classification (and especially misclassification) to identify textual sites in which to interrogate child labor, slavery and the trope of ``mere machines.''  This use of classification is not dissimilar from the interactive use of methods like topic modeling and other unsupervised techniques in the act of close reading (e.g., \citet{klein2020dimensions,walsh2021goodreads}).

\paragraph{6. Replacing human labeling at scale.}

A final use case is the application of classification as an instrumental device in downstream sensemaking: recreating human judgments at a scale beyond which people are able to carry out themselves.
After establishing the validity and interpretability of a model, \citet{long2016turbulent} then use that model as a large-scale classifier, applying it to passages from 1700 novels in order to measure the association with nationality and genre with stream-of-consciousness; in this case, classification provides a new dependent variable over a much larger scale in order to test the association of other features.
\\[10pt]
\noindent
In considering the ways in which the rise of large language models have the ability to influence this landscape of use, we might characterize those use cases by the degree to which calibrated predictive \emph{accuracy} is critical for accomplishing the intended goals, vs. the degree to which a model enables interpretative sensemaking internally. All categories rely on baseline levels of both accuracy (to validate that a model has learned from data) and sensemaking (for its use to connect to knowledge about its object of study). But we might see use case \#6 (replacing human labeling at scale) to lean more toward predictive accuracy as a goal, while use case \#1 (category sensemaking) to lean more towards internal interpretability.  We consider both of these axes in turn.   First, in such cases where accuracy alone is paramount, when are LLMs needed?  When are the tradeoffs worth it, and when do smaller, cheaper models suffice? Second, most interpretation-maximizing uses of classification in cultural analytics have generally relied on linear models (e.g., logistic regression, SVMs) to directly measure the influence of a feature on a category choice without the confounding effects of non-linearities that more complex models introduce. How are LLMs able to contribute to this goal?   To answer both of these sets of questions, we identify a set of ten datasets used in the classification categories above, and put a range of models to the test to explore this landscape.

\section{Datasets}\label{datasets}

We draw datasets from public repositories for the following papers. Most of the datasets were created to test specific hypotheses about the association of features with categories (e.g. type-token ratio with stream-of-consciousness). Across all tasks, we create standard training/development/test splits (summarized in Table \ref{tab:datasummary}) to emphasize test validity, generally decreasing the size of training data used relative to the original work.  To make the classification tasks more difficult for models, we also stratify across groups (e.g., author, genre) to isolate the target construct, so that items from the same group (e.g., texts by the same author) only appear within a single split. These modifications generally make the results reported here incomparable with the original results.

 \begin{table}[h]
     \caption{\label{tab:datasummary} 
        Dataset sizes and median number of tokens per document (in training data, tokenized through NLTK).
   }
 \small
    \centering
    \begin{tabular}{lrrrrr}
Dataset&\# train&\# dev&\# test&\# categories&median tokens\\
        \toprule

Atypical animacy&197&198&198&2&41\\
Emotion&276&276&276&8&68\\
Folktales&328&338&328&70&674\\
Genre&1250&1250&1250&5&485\\
Haiku&368&363&363&2&85\\
Hippocorpus&2285&2285&2284&3&278\\
Literary time&589&588&588&(regression)&291\\
Narrativity&4191&4841&4380&2&127\\
Strangeness&609&610&610&2&17\\
Stream-of-consciousness&200&200&200&2&246\\
        \bottomrule
    \end{tabular}   
\end{table}

\paragraph{Atypical animacy.} Data from \citet{coll-ardanuy-etal-2020-living} differentiating mentions of machine-related terms (\emph{engine}, \emph{machine}, etc.) as animate vs. non-animate.  Animate machines are those that are depicted as being alive (including people referred to as machines), while inanimate machine are not. We create train/dev/test splits at random.

\paragraph{Emotion.}  Data from \citet{Kim2018}, which identifies the emotions that characters are experiencing by tying mentions of characters in a sentence to mentions of a trigger emotion word.  We transform this into a multiclass classification task by extracting sentences in which a character experiences a single emotion, and ask a model to predict the emotion for that character. We create train/dev/test splits so that passages from the same work appear in only one partition.

\paragraph{Folktales.} Data from \citet{hagedorn2022bearing}, which assigns the ATU type\ \citep{thompson_motif_index} to texts.  We select only ATU types that are attested at least nine times among the text labels, and create train/dev/test splits to maintain the same label distribution across splits (i.e., at least three instances per type across each split).

\paragraph{Genre.}  We draw inspiration from \citet{sharmaa2020rise} in using Library of Congress subject classification as a proxy for genre, and use a subset of 5 genres that work studied (science fiction, detective and mystery stories, adventure stories, love stories, and westerns). We draw texts from Project Gutenberg, sampling 5 passages (each approximately 500 words) from 150 books for each genre.  To enable a multiclass classification problem, we only consider books that are tagged with one subject classification from the set above (so that works that are tagged with both ``love stories'' and ``westerns'' are excluded).  We create train/dev/test splits so that texts by the same author appear in only one partition, and we select a maximum of 5 books per author.

\paragraph{Haiku.} Data from \citet{long2016literary}, which contrasts haiku poems with non-haiku poems. We create train/dev/test splits so that poems by the same author appear in only one partition.

\paragraph{Hippocorpus.}  Data from \citet{sap-etal-2020-recollection}, which solicits first-person stories written by workers on Amazon Mechanical Turk in three categories: \emph{recalled} stories, which narrate real events transpiring within the past six months; \emph{imagined} stories, fictional narratives on the same topic as a randomly selected recalled story; and \emph{retold} stories, recalled stories told again 2-3 months later by the same workers.  This is the only task where a human does not judge a label by inspection of a pre-existing text; accordingly, it is not possible to articulate the textual boundaries between those categories \emph{a priori}. We create train/dev/test splits so that texts by the same author appear in only one partition.

\paragraph{Literary time.}  As our sole regression task, we draw data from Underwood\ \citep{underwood2018literary}, which labels the number of seconds that transpire in a fictional passage of approximately 250 words. We create train/dev/test splits so that texts from the same title appear in only one partition.

\paragraph{Narrativity.} Data from \citet{piper2022toward}, which contrasts passages from narrative genres (biography, fairy tales, novels) to passages from non-narrative genres (scientific abstracts, book reviews, supreme court proceedings). In this data, the genre of a text fully determines its narrativity status; in order to assess how models reason about narrativity rather than genre \emph{per se}, we create train/dev/test splits so that texts of the same genre appear in only one partition (texts of the ``fairy tale'' genre, for instance, only appear in dev data).

\paragraph{Strangeness.} Data from \citet{simeone2019towards}, which contrasts sentences mentioning ``descriptions or introductions of technology and novel science'' with those that do not, both drawn from Project Gutenberg texts.  We create train/dev/test splits at random.

\paragraph{Stream-of-consciousness.} Data from \citet{long2016turbulent}, which contrasts stream-of-consciousness passages with control passages drawn at random from realist novels.  The original work sampled control passages of fixed character lengths, leading to passages that break between words; we re-sample passages from Project Gutenberg texts of the same titles breaking only across sentences, sampling passage lengths to reflect the same empirical distribution of lengths in the SoC texts.  We create train/dev/test splits so that the passages by the same author appear in only one partition.

\section{Accuracy}

In the goals of classification for which maximizing accuracy is important, these ten tasks encode a wide range of different phenomena that shed some light on the affordances of LLMs compared to traditional supervised classifiers.  We compare the performance of the following different classes of models on these tasks.

\subsection{Supervised models}

\paragraph{Logistic/Linear regression.} We train a regularized logistic regression model for all classification tasks and ridge regression for \textsc{Literary Time}, using a bag-of-words representation of text with a vocabulary size of 100K.  We tune the $\ell_2$ regularization strength on dev data.

\paragraph{BERT/RoBERTa.}  We train base English models of BERT\ \citep{devlin-etal-2019-bert} and RoBERTa\ \citep{liu2019roberta} with a maximum 512-token input size, truncating excess text. We tune the learning rate on dev data.

\paragraph{Llama 3 8B.}  To test the ability of LLMs as supervised classifiers, we fine-tune Llama 3 8B\ \citep{llama3modelcard} on the training split with a multiclass classification head (i.e., selecting a discrete category as output, not a choice of word) for all classification tasks and a regression head for \textsc{Literary Time}, using low-rank adaptation\ \citep{hu2021lora}.  We represent a passage as the final layer output of the last word within it.  Like BERT and RoBERTa, we use a maximum 512-token input size, truncating excess text.  We tune the learning rate on dev data.
\\[10pt]
\noindent
For each task, we optimize hyperparameters on development data using the \texttt{optuna} optimization library over 50 optimization trials, training up to 100 epochs with early stopping based on lack of improvement over 10 epochs on dev data.  Appendix \ref{appendix:lr} illustrates the relationship between learning rate and accuracy on development data across all tasks for BERT, RoBERTa and Llama 3 8B; we see the best learning rate to be highly task- and model-specific, necessitating optimization.

\subsection{Prompting models}

We select three LLMs to evaluate classification through prompting alone.  For all models, we provide a description of the task and provide 10 shots as exemplars, with no Chain-of-Thought. As Appendix \ref{shots} illustrates, we experiment with shot selection and Chain-of-Thought prompting on development data with GPT-4o, finding that while the number of shots generally improves performance (though not always significantly so), using Chain-of-Thought does not lead to any meaningful difference.

\paragraph{GPT-4o.} We query GPT-4o through the OpenAI API.  Across all tasks, inference over the test set cost \$164.49.

\paragraph{Llama 3 70B.}  We run Llama 3 70B-Instruct locally across 4 L40S GPUs with a total of 192 GB of GPU memory.  The Llama 3 context size is 8,192 tokens, which is shorter than the prompt length for the \textsc{Folktales} task (which includes descriptions for the 70 categories), so we do not assess Llama 3 on this task.

\paragraph{Mixtral 8x22B.} We run Mixtral 8x22B-Instruct locally across 4 L40S GPUs with a total of 192 GB of GPU memory.  Since Mixtral models are more memory-demanding than alternatives due to their mixture of experts, we load them with 4-bit quantization.

\begin{table}[ht!]
\caption{\label{taskperf} Task performance, ordered by absolute difference between best performing prompting model and best performing supervised model. Colors visualize the proximity of that value to 1.0 (blue) and 0.0 (red).}
\scriptsize
\centering
\begin{tabular}{llrlrlrlr}
\toprule
\multicolumn{1}{c}{} & \multicolumn{2}{c}{\textbf{Folktales}} & \multicolumn{2}{c}{\textbf{Animacy}} & \multicolumn{2}{c}{\textbf{Genre}} \\
\cmidrule(lr){2-3} \cmidrule(lr){4-5} \cmidrule(lr){6-7} 
& \textbf{Accuracy} & \textbf{Time} & \textbf{Accuracy} & \textbf{Time} & \textbf{Accuracy} & \textbf{Time}  \\
\midrule
Majority & \gradient{0.033} {\tiny }& & \gradient{0.641} {\tiny }& & \gradient{0.200} {\tiny }& \\
Linear & \gradient{0.457} {\tiny [0.405-0.509]}&14.0 & \gradient{0.778} {\tiny [0.717-0.833]}&0.0 & \gradient{0.528} {\tiny [0.502-0.556]}&12.5 \\
BERT & \gradient{0.494} {\tiny [0.442-0.546]}&82.6 & \gradient{0.843} {\tiny [0.788-0.889]}&12.0 & \gradient{0.614} {\tiny [0.588-0.641]}&194.4 \\
RoBERTa & \gradient{0.616} {\tiny [0.567-0.665]}&174.4 & \gradient{0.833} {\tiny [0.783-0.884]}&10.0 & \gradient{0.648} {\tiny [0.623-0.674]}&195.0 \\
Llama 3 8B & \gradient{0.366} {\tiny [0.311-0.421]}&672.5 & \gradient{0.828} {\tiny [0.773-0.879]}&57.8 & \gradient{0.741} {\tiny [0.717-0.765]}&2013.5 \\
\midrule 
GPT-4o & \gradient{0.838} {\tiny [0.799-0.875]}& & \gradient{0.848} {\tiny [0.798-0.899]}& & \gradient{0.710} {\tiny [0.683-0.734]}& \\
Llama 3 70B & \quad\; ---& & \gradient{0.869} {\tiny [0.818-0.914]}& & \gradient{0.724} {\tiny [0.699-0.750]}& \\
Mixtral 8x22B & \gradient{0.488} {\tiny [0.433-0.543]}& & \gradient{0.823} {\tiny [0.763-0.874]}& & \gradient{0.380} {\tiny [0.351-0.406]}& \\
\bottomrule
\end{tabular}
\\[10pt]
\begin{tabular}{llrlrlrlr}
\toprule
\multicolumn{1}{c}{} & \multicolumn{2}{c}{\textbf{Stream of consciousness}} & \multicolumn{2}{c}{\textbf{Emotion}} & \multicolumn{2}{c}{\textbf{Narrativity}} \\
\cmidrule(lr){2-3} \cmidrule(lr){4-5} \cmidrule(lr){6-7}
& \textbf{Accuracy} & \textbf{Time} & \textbf{Accuracy} & \textbf{Time} & \textbf{Accuracy} & \textbf{Time} \\
\midrule
Majority & \gradient{0.500} {\tiny }& & \gradient{0.207} {\tiny }& & \gradient{0.601} {\tiny }& \\
Linear & \gradient{0.875} {\tiny [0.830-0.920]}&0.0 & \gradient{0.181} {\tiny [0.138-0.225]}&15.7 & \gradient{0.875} {\tiny [0.866-0.884]}&3.7 \\
BERT & \gradient{0.905} {\tiny [0.865-0.945]}&29.9 & \gradient{0.402} {\tiny [0.344-0.464]}&20.3 & \gradient{0.863} {\tiny [0.853-0.873]}&334.4 \\
RoBERTa & \gradient{0.900} {\tiny [0.855-0.940]}&24.6 & \gradient{0.362} {\tiny [0.308-0.417]}&22.7 & \gradient{0.928} {\tiny [0.920-0.935]}&442.0 \\
Llama 3 8B & \gradient{0.945}{\tiny [0.910-0.975]}&179.5 & \gradient{0.250} {\tiny [0.203-0.304]}&106.1 & \gradient{0.933} {\tiny [0.926-0.940]}&2635.2 \\
\midrule
GPT-4o & \gradient{0.925} {\tiny [0.885-0.960]}& & \gradient{0.370} {\tiny [0.315-0.428]}& & \gradient{0.830} {\tiny [0.818-0.840]}& \\
Llama 3 70B & \gradient{0.875} {\tiny [0.825-0.920]}& & \gradient{0.373} {\tiny [0.312-0.435]}& & \gradient{0.875} {\tiny [0.865-0.884]}& \\
Mixtral 8x22B & \gradient{0.820} {\tiny [0.765-0.870]}& & \gradient{0.163} {\tiny [0.123-0.207]}& & \gradient{0.604} {\tiny [0.591-0.618]}& \\
\bottomrule
\end{tabular}
\\[10pt]
\begin{tabular}{llrlrlrlr}
\toprule
\multicolumn{1}{c}{} & \multicolumn{2}{c}{\textbf{Strangeness}} & \multicolumn{2}{c}{\textbf{Haiku}} & \multicolumn{2}{c}{\textbf{Hippocorpus}} \\
\cmidrule(lr){2-3} \cmidrule(lr){4-5} \cmidrule(lr){6-7}
& \textbf{Accuracy} & \textbf{Time} & \textbf{Accuracy} & \textbf{Time} & \textbf{Accuracy} & \textbf{Time} \\
\midrule
Majority & \gradient{0.556} {\tiny }& & \gradient{0.529} {\tiny }& & \gradient{0.405} {\tiny }& \\
Linear & \gradient{0.805} {\tiny [0.772-0.834]}&0.0 & \gradient{0.705} {\tiny [0.658-0.752]}&0.0 & \gradient{0.518} {\tiny [0.497-0.539]}&24.2 \\
BERT & \gradient{0.879} {\tiny [0.852-0.903]}&17.9 & \gradient{0.782} {\tiny [0.738-0.824]}&34.0 & \gradient{0.615} {\tiny [0.598-0.636]}&289.7 \\
RoBERTa & \gradient{0.877} {\tiny [0.849-0.903]}&23.9 & \gradient{0.986} {\tiny [0.972-0.997]}&56.9 & \gradient{0.646} {\tiny [0.624-0.665]}&511.0 \\
Llama 3 8B & \gradient{0.874} {\tiny [0.848-0.900]}&145.8 & \gradient{0.992} {\tiny [0.981-1.000]}&296.0 & \gradient{0.682} {\tiny [0.662-0.699]}&4228.8 \\
\midrule
GPT-4o & \gradient{0.564} {\tiny [0.523-0.602]}& & \gradient{0.785} {\tiny [0.741-0.826]}& & \gradient{0.408} {\tiny [0.388-0.429]}& \\
Llama 3 70B & \gradient{0.726} {\tiny [0.690-0.761]}& & \gradient{0.625} {\tiny [0.576-0.675]}& & \gradient{0.412} {\tiny [0.391-0.432]}& \\
Mixtral 8x22B & \gradient{0.761} {\tiny [0.726-0.793]}& & \gradient{0.813} {\tiny [0.771-0.851]}& & \gradient{0.263} {\tiny [0.245-0.279]}& \\
\bottomrule
\end{tabular}
\\[10pt]
\begin{tabular}{llrlrlrlr}
\toprule
\multicolumn{1}{c}{} & \multicolumn{2}{c}{\textbf{Literary Time}} \\
\cmidrule(lr){2-3}
& \textbf{$\rho$} & \textbf{Time} \\
\midrule
Majority & \gradient{0} {\tiny }& \\
Linear & \gradient{0.640} {\tiny [0.588-0.687]}&0.2 \\
BERT & \gradient{0.764} {\tiny [0.720-0.805]}&88.6 \\
RoBERTa & \gradient{0.782} {\tiny [0.738-0.817]}&111.4 \\
Llama 3 8B & \gradient{0.772} {\tiny [0.730-0.809]}&966.3 \\
\midrule 
GPT-4o & \gradient{0.485} {\tiny [0.406-0.560]}& \\
Llama 3 70B & \gradient{0.371} {\tiny [0.285-0.455]}& \\
Mixtral 8x22B & \gradient{0.447} {\tiny [0.378-0.508]}& \\
\bottomrule
\end{tabular}
\end{table}

\subsection{Discussion}
We present results in Table \ref{taskperf}, reporting accuracy as a metric for all classification tasks and Spearman $\rho$ for \textsc{Literary time}.  For all supervised models, we report the total wall clock training time in minutes across all hyperparameter optimization trials.  Tasks are ordered (by left to right and top to bottom) by the difference in metric between best performing prompting method and best supervised method.

\paragraph{Established vs. \emph{de novo} tasks.} The diversity of tasks on which we assess models allows us to characterize disparate performance among them. Tasks measuring concepts that are generally widely known---the animacy of objects, determinations of science fiction vs. westerns, folktale types, character emotional states, and stream-of-consciousness---offer sites where LLMs are able to excel, either beating supervised models or coming close in performance to them.  For tasks that measure largely \emph{de novo} concepts---the passage of time, differentiating stories that are recalled from those that are imagined or retold, ``technological strangeness''---the evidence from supervision appears to be important for accurate prediction.

\paragraph{Memorization.}

The outsized performance of GPT-4o on the folktale identification task raises the question whether that model has memorized that task; as we detail in Appendix \ref{app:memorization}, we assess the influence of memorization on task performance using a membership inference method inspired by \citet{shi2024detecting}.  We see little evidence of memorization, and after applying correction for multiple hypothesis correction, no evidence for downstream impact.

\paragraph{Sufficiency of masked LMs.}  Among models that are optimized for each specific task using supervised data, we do not see a strong difference between base masked language models and Llama 3, which is pre-trained on a much larger amount of data, has 65 times more parameters, and generally takes an order of magnitude longer to train.

\section{Sensemaking}

The comparative performance we see reported above only speaks to high classification accuracy as a desideratum.  This leaves aside the other, perhaps more important, ways in which we see classification being used as a tool for sensemaking in cultural analytics.

If we probe deeper into the ways in which linear classification methods have been used for making sense of categories,  both theory-driven approaches and exploratory analysis make use of \emph{conceptualization} (deciding on specific constructs to measure) and \emph{implementation} (instantiating those constructs into specific measures), to use the terminology from \citet{piper2017think}.  A primary difference, however, comes in the role of prior knowledge and \emph{theorization} guiding those choices.  Top-down approaches often design models to test theories; exploratory approaches may also incorporate theory in the design of the representation space $f(x)$, but a starting point is often to let the structure of the data guide our understanding of the concepts themselves. Topic modeling is a widely used example of this form; an unsupervised method learns structure from data which then guides the analysis that follows.  We see many examples of this approach. \ \citet{goldstone2014quiet} use topic modeling to first learn that ``critical theory'' is a dominant theme in literary journals; after surfacing this fact, they more formally test its rise and fall over time; \citet{bode2018} likewise uses topic modeling to identify a set of themes organizing a collection of Australian, British and American fiction, then uses close reading and classification to narrow in on a formal analysis of nonmetropolitan colonial spaces as a major organizing principle.  Sensemaking as an activity is driven in these cases by the structure found through exploratory models. Indeed, the practice of computational grounded theory\ \citep{nelson2020computational} is premised on exactly this movement between exploration through models, close reading, and formal testing.

If we consider the ways in which LLMs are able to serve some function in this larger sensemaking process, a natural starting point is their role in exploratory analysis.  What structure exists in the data that can lead to formal theory-testing?  We experiment with this use case by prompting GPT-4o to reason about the characteristics that differentiate a set of data points attended by anonymized category labels, as illustrated in Figure \ref{catprompt}.  We include as many data points as the context length allows.

\begin{figure}
    \centering
\begin{tcolorbox}[colframe=black,colback=white,boxrule=0.5mm,arc=0mm]
Consider the data below, which contains a list of text/label pairs that illustrate a set of categories (0, 1):
\\[10pt]
	[\{"text": "Strong called to the remaining ships of his fleet Take formation K. Land and attack the enemy base according to prearranged order.", "label": 0\}\\
    \{"text": "She dithered for a moment then stepped back toward the kitchen and the stairwell.", "label": 1\}

    \ldots ]

	Using this data, and this data alone, what are the textual characteristics that differentiate the classes from each other?  Provide a list of bullet points of those textual features in the following format:
\\[10pt]
	\#\#\# Category 0 Characteristics:

	- **\\

	\#\#\# Category 1 Characteristics:

	- **
\end{tcolorbox}
    \caption{Template for LLM sensemaking exercise, with anonymized integer labels, illustrated with data for the \textsc{Strangeness} task (0 = \emph{strange}; 1 = \emph{not strange}).\label{catprompt} }
\end{figure}

\begin{table}
  \caption{GPT-4o output for characterizing  the difference between Haiku and Non-Haiku poems through anonymized labels (only presented to the model as class "1" vs. "0"); for brevity, the first 5 characteristics are presented here; the full list can be found in Appendix \ref{gpt:haiku}.}
  \label{tab:gpt4ohaiku}
  \centering
  \begin{tabular}{p{7cm}p{7cm}}
    \toprule
    Haiku & Non-Haiku \\
    \midrule
\textbf{Concise and Lyrical}: Texts are often shorter, more concise, and have a lyrical quality. & \textbf{Narrative or Descriptive Style}: Texts often tell a story or describe a scene in detail.\\

\textbf{Nature Imagery}: Frequent use of natural imagery, such as flowers, trees, and weather. &\textbf{Complex Structure}: Sentences are often longer and more complex, with multiple clauses.
 \\

\textbf{Emotional and Sensory Language}: Focus on emotions and sensory experiences. &\textbf{Historical or Mythological References}: Frequent references to historical events, mythological figures, or classical literature.
 \\

\textbf{Simple Structure}: Sentences are generally simpler and more straightforward. &\textbf{Character Focus}: Emphasis on characters, their actions, and their emotions.
 \\

\textbf{Personal and Intimate}: Often feels personal or intimate, as if sharing a private moment or thought. &\textbf{Dialogue and Monologue}: Presence of dialogue or internal monologue.
 \\

    \bottomrule
  \end{tabular}
\end{table}

Table \ref{tab:gpt4ohaiku} presents the output of this experiment for the \textsc{Haiku} task; the full outputs for all ten tasks can be found in Appendix \ref{llmsensemaking}.  Here we see similar affordances for sensemaking as alternative models. Just as  \citet{long2016literary} challenge the initial category boundaries of haiku by identifying naturalistic language and brevity as defining characteristics of it, so too does GPT-4o identify ``nature imagery'' and conciseness as characteristic.  An examination of the outputs in Appendix \ref{llmsensemaking} shows that the sensemaking enabled by this form of LLM prompting intersects with that enabled by alternative methods in these tasks' original papers. The defining features of \textsc{Animacy} identified by GPT-4o stress ``the mechanical nature of humans or the human-like qualities of machines,'' (\S \ref{gptanimacy}) including the focus identified by \citet{wilson2023living} on ``mere machines.'' This method identifies ``narrative style'' vs. ``lack of narrative'' (\S \ref{gpt:narrativity}) as the defining feature of the \textsc{Narrativity} task but offers alternatives to the distant worlds theory of \citet{piper2022toward}---offering ``descriptive language'' and ``imagery'' as additional characteristics.  In many ways, prompting LLMs to find the ways in which a set of categories are differentiated from each other within the evidence of a labeled dataset can suggest meaningful directions for more formal interrogation.

At the same time, it is clear that such LLM characterizations are not to be trusted at face value; we know this from prior work on LLM hallucinations\ \citep{huang2023survey,xu2024hallucination,kalai2024calibrated} and confabulations\ \citep{sui2024confabulation}, but also in examining its points of failure within this set of tasks. For \textsc{Emotion} (\S \ref{gpt:emotion}), we see clear evidence of overfitting (e.g., ``references to royalty or nobility'' for the anticipation class, ``mentions of specific names'' for joy); and its description of the \textsc{Folktale} task (\S\ref{gpt:folktales}) fails to meaningfully differentiate categories that it performs well on when given their description.  Additionally, one of the risks of using pre-trained LLMs for exploratory data analysis is the degree to which the analysis they drive is a direct result of examining the data given, or a function instead of task knowledge acquired during pre-training. While we are careful to anonymize the category labels in the experiments above and not mention the classification task by name (e.g., ``haiku''), it is clear that pre-training data informs the knowledge that models bring to this exploratory task, muddying the inferences we can make about data-in-itself.  Understanding the role that pre-training plays both in model performance (e.g., through disparate memorization) and in any biases that result (e.g., through disparate focus on whose language is represented\ \cite{lucy-etal-2024-aboutme}) is an important area of current research on the limits of these models.

\subsection{Discussion}

As an exploratory exercise, the defining characteristics of categories seen through the lens of a prompted LLM provide suggestions on potentially salient dimensions that only later formal testing can truly assess the significance of.  Is it indeed the case that westerns ``focus on personal experiences and emotions'' more than other genres (\S\ref{gpt:genre})?  Is sentence length indeed correlated with the duration of elapsed time (\S\ref{gpt:literarytime})?  While not providing an answer themselves, they suggest potentially interesting directions to test with more formal means (i.e., put through the stages of {conceptualization} and {implementation}).  In this sense, we might view LLMs as tools for sensemaking---helping us understand categories and the boundaries that circumscribe them---in the same way that other exploratory methods like topic modeling make possible.

\section{Conclusion}

In this work, we outline the major ways in which current work in cultural analytics is employing \emph{classification}, primarily as a method for sensemaking distinct from mere automation, and detail the ways in which large language models fit (and do not fit) within this landscape.  For uses that seek to maximize predictive accuracy, LLMs offer competitive performance through prompting alone for established tasks, while traditional supervised methods excel for newly constructed phenomena (even in scenarios with limited training data).  For sensemaking applications that aim to understand category constructs better through labeled data, LLMs may have a role to play in exploratory data analysis by outlining potential characteristics that can then be subjected to more formal testing.  
By considering a set of ten tasks considered in the space of cultural analytics, we complement an existing body of work that probes model knowledge of culture\ \citep{liu-etal-2024-multilingual,shen-etal-2024-understanding}) to focus in particular on the ways in which computational models can shed light on culture itself.  
Data and code to support this work can be found at \url{https://github.com/bamman-group/ca-classification-data}.

\begin{acknowledgments}
    We thank the original authors of the tasks we study for openly releasing the data on which this current work stands.
    The research reported in this article was supported by funding from the National Science Foundation (IIS-1942591) and the National Endowment for the Humanities (HAA-271654-20), with computing resources provided by Microsoft Azure (Accelerating Foundation Models Research) and the Accelerating Computing for Emerging Sciences (ACES) high-performance computing cluster at Texas A\&M University, funded by the National Science Foundation (NAIRR-240114).

\end{acknowledgments}
\clearpage

\bibliography{bibliography}

\section*{Appendices}

\appendix

\section{Shots and Chain-of-Thought}\label{shots}

We investigate the relationship between number of shots, Chain-of-Thought reasoning and accuracy for GPT-4o in Table \ref{tab:shots} below.

 \begin{table}[h]
     \caption{\label{tab:shots} 
        GPT-4o dev performance by number of shots/chain of thought.  Increasing the number of shots (from 1 to 10) largely helps performance (e.g., for animacy, genre, literary time, and strangeness), even if not significantly so.  CoT generally does not improve performance.  We do not test CoT on \textsc{Hippocorpus} since it does not originate as a human labeling task (as noted in \S\ref{datasets} above).
   }
 \small
    \centering

    \begin{tabular}{lccc}
Task&Shots&No CoT&CoT\\
        \toprule
Animacy&1&0.773 [0.712-0.828]&0.707 [0.641-0.768]\\
&5&0.833 [0.778-0.884]&0.783 [0.727-0.838]\\
&10&0.828 [0.778-0.879]&0.833 [0.783-0.884]\\
\midrule
Emotion&1&0.362 [0.304-0.417]&0.391 [0.333-0.446]\\
&5&0.384 [0.330-0.442]&0.420 [0.362-0.482]\\
&10&0.384 [0.322-0.442]&0.395 [0.337-0.453]\\
\midrule
Folktales&1&0.851 [0.811-0.887]&0.872 [0.835-0.905]\\
&5&0.854 [0.811-0.890]&0.860 [0.823-0.896]\\
&10&0.854 [0.814-0.890]&0.857 [0.817-0.893]\\
\midrule
Genre&1&0.674 [0.649-0.699]&0.674 [0.646-0.699]\\
&5&0.693 [0.669-0.719]&0.682 [0.658-0.707]\\
&10&0.729 [0.704-0.752]&0.682 [0.657-0.709]\\
\midrule
Haiku&1&0.766 [0.724-0.807]&0.711 [0.661-0.752]\\
&5&0.793 [0.749-0.832]&0.747 [0.700-0.793]\\
&10&0.777 [0.733-0.818]&0.747 [0.700-0.791]\\
\midrule
Hippocorpus&1&0.404 [0.384-0.424]&\\
&5&0.399 [0.381-0.419]&\\
&10&0.400 [0.379-0.420]&\\
\midrule
Literary time&1&0.231 [0.146-0.323]&0.249 [0.154-0.339]\\
&5&0.490 [0.411-0.569]&0.498 [0.415-0.567]\\
&10&0.508 [0.433-0.576]&0.489 [0.409-0.570]\\
\midrule
Narrativity&1&{0.885} [0.876-0.895]&{0.899} [0.890-0.907]\\
&5&0.814 [0.803-0.825]&0.850 [0.839-0.860]\\
&10&0.820 [0.809-0.832]&0.862 [0.852-0.872]\\
\midrule
Strangeness&1&0.474 [0.434-0.515]&0.510 [0.469-0.546]\\
&5&0.544 [0.505-0.584]&0.575 [0.539-0.615]\\
&10&0.582 [0.544-0.621]&0.584 [0.544-0.623]\\
\midrule
Stream-of-consciousness&1&0.910 [0.870-0.950]&0.905 [0.860-0.945]\\
&5&0.890 [0.845-0.930]&0.845 [0.795-0.895]\\
&10&0.895 [0.850-0.935]&0.865 [0.815-0.910]\\
        \bottomrule
    \end{tabular}
\end{table}

\clearpage
\section{Memorization}\label{app:memorization}

The outsized performance of GPT-4o on the folktale identification task raises the question whether that model has memorized that task; indeed, we can see that the text itself is often discoverable on Google attended with its label, as illustrated in Figure \ref{google} below.

\begin{figure}[h]
    \centering
    \includegraphics[width=.7\linewidth]{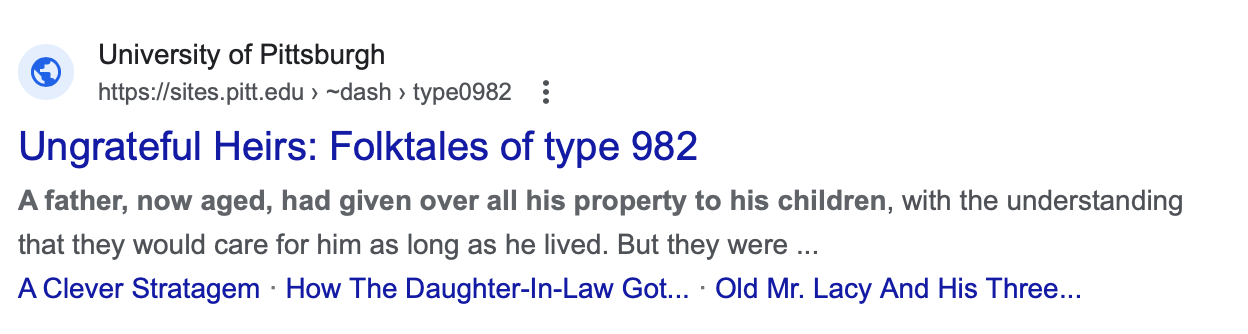}
    \caption{Searching for the text of a data point in the \textsc{Folktales} training set (``A father, now aged, had given over all his property to his children'') brings up its label (982).}
    \label{google}
\end{figure}

Prior work in cultural analytics has assessed the impact of memorization on downstream tasks, with work demonstrating memorization influence on encyclopedic knowledge (predicting date of publication\ \citep{chang-etal-2023-speak}) but not on text-internal qualities like poetic form\ \cite{walsh2024sonnet}.

We assess instance-level memorization in prompt-based LLMs using a membership inference method similar to \citet{shi2024detecting}; while that work identifies membership through the log probabilities of the $k$ least probable tokens in a text (necessitating probability assessments for every token), we identify first the $k=5$ least probable tokens (relative to the complete data for that task) and ask a model to predict them given a fixed context length of 25 tokens, as in the example below:

\begin{center}
\fbox{\parbox{10cm}{
    My wife and I have been living in the same family home for the last [MASK] years . It was our first family home and $\rightarrow$ \textbf{ten}
}}
\end{center}
We judge a passage to be memorized if at least four of the five masked terms are able to be predicted. We measure the influence of memorization on the task through the Pearson correlation between an instance's memorization and whether the model correctly predicted its label (in the case of regression for \textsc{Literary Time}, the mean absolute error between true and predicted label); for correlation between two binary variables, this is equivalent to the $\phi$ coefficient\ \citep{yule1912methods}.   We select this method over others due to the specific nature of these datasets, which do not always contain names (required by \citet{chang-etal-2023-speak}) and do not always have titles or authors notable enough to condition on (required by \citet{d2023chatbot}).

 As Table \ref{memtable} shows, while  GPT-4o on \textsc{Folktales}, \textsc{hippocorpus} and \textsc{literary time} all rank highest on memorization, there is no measurable effect on downstream tasks for any model/task combination when correcting for multiple hypothesis tests.

This assessment of memorization, however, only considers memorization of the text alone, and not the association of text with a label.  While past work has examined this form of label memorization in controlled settings where the training data is known\ \citep{magar-schwartz-2022-data}, it remains a difficult task for membership \emph{inference}---and in fact may be impossible to disentangle from the understanding of the task itself.

\begin{table}[ht]
\scriptsize
\centering
\caption{\label{memtable} Dataset memorization by prompting models.  With Bonferroni correction for the $n=29$ hypothesis tests carried out, no model shows significant correlation at $\alpha=0.05$ between memorization and task performance.  We do not assess \textsc{Folktales} on Llama 3 70B since the prompt for that task exceeds the Llama context size (as noted above).  For cases with 0 memorization, Pearson's $r$ is undefined (denoted by blank values).}
\begin{tabular}{cccccccccc}
\toprule
\multicolumn{1}{c}{} & \multicolumn{3}{c}{\textbf{GPT-4o}} & \multicolumn{3}{c}{\textbf{Llama 3 70B}} & \multicolumn{3}{c}{\textbf{Mixtral 8x22B}} \\
\cmidrule(lr){2-4} \cmidrule(lr){5-7} \cmidrule(lr){8-10} 
& \textbf{Mem} & \textbf{$r$} & \textbf{$p$} & \textbf{Mem} & \textbf{$r$} & \textbf{$p$}& \textbf{Mem} & \textbf{$r$} & \textbf{$p$}  \\
\midrule
Animacy & 0.010&-0.098&0.169 & 0.000&& & 0.000&& \\
Emotion & 0.022&-0.063&0.300 & 0.007&-0.066&0.271 & 0.004&-0.027&0.660 \\
Folktales & 0.116&0.027&0.624 & ---&& & 0.006&0.002&0.976 \\
Genre & 0.022&-0.022&0.432 & 0.010&-0.060&0.034 & 0.002&-0.005&0.868 \\
Haiku & 0.003&0.027&0.602 & 0.003&0.041&0.440 & 0.000&& \\
Hippocorpus & 0.137&-0.021&0.327 & 0.067&0.005&0.814 & 0.006&0.048&0.023 \\
Literary time & 0.034&0.001&0.990 & 0.003&-0.013&0.744 & 0.000&& \\
Narrativity & 0.076&0.042&0.005 & 0.032&0.037&0.014 & 0.005&0.020&0.186 \\
Strangeness & 0.040&-0.039&0.341 & 0.015&0.016&0.703 & 0.000&& \\
Stream-of-consciousness & 0.030&0.050&0.481 & 0.005&0.027&0.706 & 0.000&& \\
\bottomrule
\end{tabular}
\end{table}

\clearpage
\section{LLMs for sensemaking}\label{llmsensemaking}

\subsection{Animacy}\label{gptanimacy}

\noindent Category 0 (sc. animate) Characteristics:
\begin{enumerate}
\item \textbf{Human or animal comparison}: Texts often compare machines to humans or animals, emphasizing the mechanical nature of humans or the human-like qualities of machines (e.g., ``servants must be treated, not as machines, but as human beings'').

\item \textbf{Metaphorical or figurative language}: Usage of machines in a metaphorical sense to describe people or actions (e.g., ``a mere machine'', ``a teaching machine'').

\item \textbf{Negative connotation}: Machines are often described in a negative or dehumanizing context (e.g., ``mere machines'', ``living machine'').

\item \textbf{Lack of technical detail}: Texts do not provide specific technical details about the machines or their operations.

\item \textbf{Focus on human experience}: Emphasis on human emotions, experiences, or societal roles in relation to machines (e.g., ``a soldier is, of all descriptions of men, the most completely a machine'').

\end{enumerate}
\noindent Category 1 (sc. non-animate) Characteristics:
\begin{enumerate}
\item \textbf{Technical or operational context}: Texts often describe machines in a technical or operational context, detailing their functions or uses (e.g., ``printing machines of automatic construction'', ``engines employed in pumping'').

\item \textbf{Specific machine types}: Mention of specific types of machines or apparatus (e.g., ``steam engine'', ``locomotive'', ``dynamo'').

\item \textbf{Positive or neutral connotation}: Machines are described in a functional, neutral, or positive light, often highlighting their utility or innovation (e.g., ``valuable improvement on Lamb's knitting machine'', ``self-feeders'').

\item \textbf{Historical or factual description}: Texts provide historical or factual descriptions of machines and their development or usage (e.g., ``The first steam engine for grinding burnt flint'').

\item \textbf{Detailed descriptions}: Detailed descriptions of the machines' components, operations, or the context in which they are used (e.g., ``The drum of the King pit engine on the Wigan Four-feet rope side is 15 feet 2 inches in diameter'').

\end{enumerate}

\subsection{Emotion}\label{gpt:emotion}

\noindent Category 0 (sc. anticipation) Characteristics:
\begin{enumerate}
\item \textbf{References to royalty or nobility}: Mentions of princes, kings, or other noble titles.

\item \textbf{Dialogue involving commands or requests}: Characters giving orders or making requests.

\item \textbf{Descriptions of physical appearance}: Comments on someone's looks, such as ``beautiful boots.''

\end{enumerate}
\noindent Category 1 (sc. disgust) Characteristics:
\begin{enumerate}
\item \textbf{Philosophical or reflective tone}: Discussions about life, existence, or deep thoughts.

\item \textbf{Expressions of doubt or uncertainty}: Characters questioning or reflecting on their actions or beliefs.

\item \textbf{Mentions of emotions like sadness or irritation}: Characters expressing feelings of melancholy or frustration.

\end{enumerate}
\noindent Category 2 (sc. joy) Characteristics:
\begin{enumerate}
\item \textbf{Mentions of specific names}: Characters like Dora, Miss Barfoot, or Miss Stackpole.

\item \textbf{Dialogue involving personal relationships}: Conversations about friends, family, or acquaintances.

\item \textbf{References to beauty or physical appearance}: Comments on someone's looks or attractiveness.

\end{enumerate}
\noindent Category 3 (sc. sadness) Characteristics:
\begin{enumerate}
\item \textbf{Mentions of groups or collective actions}: References to groups of people, such as ``the crowd'' or ``the little garrison.''

\item \textbf{Descriptions of conflict or tension}: Situations involving arguments, fights, or other forms of conflict.

\item \textbf{References to specific locations or settings}: Mentions of places like military prisons or specific rooms.

\end{enumerate}
\noindent Category 4 (sc. anger) Characteristics:
\begin{enumerate}
\item \textbf{Mentions of nature or the environment}: Descriptions of natural settings, weather, or animals.

\item \textbf{Expressions of fear or danger}: Characters feeling threatened or in peril.

\item \textbf{References to physical sensations or conditions}: Descriptions of physical states, such as being cold or in pain.

\end{enumerate}
\noindent Category 5 (sc. surprise) Characteristics:
\begin{enumerate}
\item \textbf{Mentions of family or domestic settings}: References to family members or home life.

\item \textbf{Expressions of surprise or disbelief}: Characters reacting with astonishment or incredulity.

\item \textbf{References to specific actions or events}: Descriptions of particular incidents or activities.

\end{enumerate}
\noindent Category 6 (sc. fear) Characteristics:
\begin{enumerate}
\item \textbf{Mentions of specific names}: Characters like Sister Martha, Henriot, or Charles.

\item \textbf{Descriptions of medical or health-related situations}: References to doctors, patients, or health conditions.

\item \textbf{Expressions of fear or danger}: Characters feeling threatened or in peril.

\end{enumerate}
\noindent Category 7 (sc. trust) Characteristics:
\begin{enumerate}
\item \textbf{Dialogue involving personal relationships}: Conversations about friends, family, or acquaintances.

\item \textbf{Expressions of emotions like sadness or happiness}: Characters expressing feelings of joy or sorrow.

\item \textbf{References to specific actions or events}: Descriptions of particular incidents or activities.

\end{enumerate}

\subsection{Folktales}\label{gpt:folktales}

[Note: for legibility, we present the first 10 categories; all results can be viewed in our Github repository.]
\\[10pt]
\noindent Category 0 (sc. 982) Characteristics:
\begin{enumerate}
\item \textbf{Themes of deception and trickery}: The story involves a father deceiving his children by pretending to have money.

\item \textbf{Moral lesson}: The story teaches a lesson about gratitude and the consequences of greed.

\end{enumerate}
\noindent Category 1 (sc. 650A) Characteristics:
\begin{enumerate}
\item \textbf{Themes of laziness and cleverness}: The protagonist, Jack, is initially lazy but uses cleverness to achieve his goals.

\item \textbf{Transformation and success}: Jack transforms from a lazy boy to a successful individual through a series of clever actions.

\end{enumerate}
\noindent Category 2 (sc. 510B) Characteristics:
\begin{enumerate}
\item \textbf{Themes of enchantment and transformation}: Characters are often transformed by magical means, such as a girl being turned into a reindeer.

\item \textbf{Quest for identity and love}: The stories often involve a quest to regain one's true form or to find true love.

\end{enumerate}
\noindent Category 3 (sc. 113A) Characteristics:
\begin{enumerate}
\item \textbf{Themes of supernatural intervention}: Supernatural beings like trolls or witches play a significant role.

\item \textbf{Rural and pastoral settings}: The stories often take place in rural settings, involving common folk and simple tasks.

\end{enumerate}
\noindent Category 4 (sc. 700) Characteristics:
\begin{enumerate}
\item \textbf{Themes of cleverness and resourcefulness}: The protagonist uses cleverness to outwit others, such as a tinker finding treasure.

\item \textbf{Elements of folklore and superstition}: The stories often involve elements of folklore, such as dreams leading to treasure.

\end{enumerate}
\noindent Category 5 (sc. 175) Characteristics:
\begin{enumerate}
\item \textbf{Themes of trickery and cunning}: Characters like Brer Rabbit use cunning to outwit stronger opponents.

\item \textbf{Animal characters with human traits}: The stories often feature animals with human-like intelligence and behavior.

\end{enumerate}
\noindent Category 6 (sc. 980) Characteristics:
\begin{enumerate}
\item \textbf{Themes of family and respect}: The story involves family dynamics and the importance of respecting elders.

\item \textbf{Moral lesson}: The story teaches a lesson about treating family members with kindness and respect.

\end{enumerate}
\noindent Category 7 (sc. 726) Characteristics:
\begin{enumerate}
\item \textbf{Themes of supernatural and magical intervention}: Supernatural beings like trolls or witches play a significant role.

\item \textbf{Rural and pastoral settings}: The stories often take place in rural settings, involving common folk and simple tasks.

\end{enumerate}
\noindent Category 8 (sc. 1430) Characteristics:
\begin{enumerate}
\item \textbf{Themes of laziness and cleverness}: The protagonist, Heinz, is initially lazy but uses cleverness to achieve his goals.

\item \textbf{Transformation and success}: Heinz transforms from a lazy boy to a successful individual through a series of clever actions.

\end{enumerate}
\noindent Category 9 (sc. 2250) Characteristics:
\begin{enumerate}
\item \textbf{Themes of brevity and simplicity}: The story is very short and simple, often involving a single event or action.

\item \textbf{Lack of detailed plot}: The story lacks a detailed plot or character development.

\end{enumerate}

\subsection{Genre}\label{gpt:genre}

\noindent Category 0 (sc. science fiction) Characteristics:
\begin{enumerate}
\item \textbf{Frequent use of dialogue and conversational tone}: The texts often include characters speaking to each other, which makes the narrative dynamic and engaging.

\item \textbf{Presence of action and movement}: The texts often describe characters performing actions, moving from one place to another, or engaging in some form of physical activity.

\item \textbf{Descriptive language for settings and events}: The texts provide detailed descriptions of the environment, events, and actions, creating a vivid picture for the reader.

\item \textbf{Suspense and tension}: The texts often build suspense and tension, keeping the reader engaged and curious about what will happen next.

\end{enumerate}
\noindent Category 1 (sc. western stories) Characteristics:
\begin{enumerate}
\item \textbf{Focus on personal experiences and emotions}: The texts often delve into the personal thoughts, feelings, and experiences of the characters.

\item \textbf{Introspective and reflective tone}: The narrative often includes characters reflecting on their past actions, decisions, and emotions.

\item \textbf{Detailed character development}: The texts provide in-depth descriptions of characters' personalities, backgrounds, and motivations.

\item \textbf{Use of first-person perspective}: Some texts are written from a first-person point of view, giving a direct insight into the character's mind.

\end{enumerate}
\noindent Category 2 (sc. detective and mystery stories) Characteristics:
\begin{enumerate}
\item \textbf{Mystery and investigation elements}: The texts often involve characters trying to solve a mystery or uncover hidden information.

\item \textbf{Presence of law enforcement or detective figures}: Characters such as detectives, police officers, or investigators are commonly featured.

\item \textbf{Clues and evidence}: The narrative includes details about clues, evidence, and logical deductions made by the characters.

\item \textbf{Formal and procedural language}: The texts often use formal language and describe procedural actions related to investigations or legal matters.

\end{enumerate}
\noindent Category 3 (sc. adventure stories) Characteristics:
\begin{enumerate}
\item \textbf{Historical or period settings}: The texts are often set in historical or period-specific contexts, with detailed descriptions of the time period.

\item \textbf{Focus on social and cultural aspects}: The narrative often explores social norms, cultural practices, and historical events relevant to the time period.

\item \textbf{Descriptive language for attire and environment}: The texts provide detailed descriptions of clothing, architecture, and other elements specific to the historical setting.

\item \textbf{Formal and elaborate language}: The language used is often more formal and elaborate, reflecting the historical context.

\end{enumerate}
\noindent Category 4 (sc. love stories) Characteristics:
\begin{enumerate}
\item \textbf{Romantic and emotional themes}: The texts often focus on romantic relationships, emotional experiences, and personal connections between characters.

\item \textbf{Descriptive language for emotions and interactions}: The narrative includes detailed descriptions of characters' emotions, facial expressions, and interactions.

\item \textbf{Use of dialogue to convey feelings}: Dialogue is often used to express characters' feelings, thoughts, and relationship dynamics.

\item \textbf{Focus on personal and intimate moments}: The texts highlight personal and intimate moments between characters, emphasizing their emotional significance.

\end{enumerate}

\subsection{Haiku}\label{gpt:haiku}

\noindent Category 0 (sc. non-haiku) Characteristics:
\begin{enumerate}
\item \textbf{Narrative or Descriptive Style}: Texts often tell a story or describe a scene in detail.

\item \textbf{Complex Structure}: Sentences are often longer and more complex, with multiple clauses.

\item \textbf{Historical or Mythological References}: Frequent references to historical events, mythological figures, or classical literature.

\item \textbf{Character Focus}: Emphasis on characters, their actions, and their emotions.

\item \textbf{Dialogue and Monologue}: Presence of dialogue or internal monologue.

\item \textbf{Detailed Imagery}: Rich, detailed imagery that paints a vivid picture.

\item \textbf{Formal Tone}: Often a more formal or archaic tone.

\item \textbf{Philosophical or Reflective}: Texts may include philosophical musings or reflections on life and existence.

\item \textbf{Narrative Progression}: Clear progression of events or thoughts.

\item \textbf{Use of Direct Address}: Often addresses the reader or another character directly.

\end{enumerate}
\noindent Category 1 (sc. haiku) Characteristics:
\begin{enumerate}
\item \textbf{Concise and Lyrical}: Texts are often shorter, more concise, and have a lyrical quality.

\item \textbf{Nature Imagery}: Frequent use of natural imagery, such as flowers, trees, and weather.

\item \textbf{Emotional and Sensory Language}: Focus on emotions and sensory experiences.

\item \textbf{Simple Structure}: Sentences are generally simpler and more straightforward.

\item \textbf{Personal and Intimate}: Often feels personal or intimate, as if sharing a private moment or thought.

\item \textbf{Symbolism and Metaphor}: Heavy use of symbolism and metaphor to convey deeper meanings.

\item \textbf{Reflective and Meditative}: Texts often reflect on a single idea or moment.

\item \textbf{Visual and Sensory Details}: Emphasis on visual and sensory details to create a mood or atmosphere.

\item \textbf{Ephemeral and Transient Themes}: Themes of impermanence, change, and fleeting moments.

\item \textbf{Minimalist Style}: A minimalist approach with fewer words and more impact.

\end{enumerate}

\subsection{Hippocorpus}

\noindent Category 0 (sc. retold) Characteristics:
\begin{enumerate}
\item \textbf{Focus on specific events or actions:} Entries in this category often describe a particular event or action in detail, such as a specific activity, task, or incident.

\item \textbf{Lack of emotional depth:} These entries tend to be more factual and less emotionally charged, focusing on the sequence of events rather than the emotional impact.

\item \textbf{Shorter and more concise:} The descriptions are often brief and to the point, without extensive elaboration or reflection.

\item \textbf{Less personal reflection:} There is minimal introspection or personal reflection on the significance of the event.

\end{enumerate}
\noindent Category 1 (sc. imagined) Characteristics:
\begin{enumerate}
\item \textbf{Personal achievements or milestones:} Entries often describe personal accomplishments, milestones, or significant life changes.

\item \textbf{Positive or neutral tone:} The tone is generally positive or neutral, focusing on achievements, celebrations, or positive changes.

\item \textbf{Detailed descriptions:} These entries provide detailed descriptions of events, including specific actions taken and their outcomes.

\item \textbf{Emotional but not overly dramatic:} While there is some emotional content, it is not as intense or dramatic as in Category 2.

\end{enumerate}
\noindent Category 2 (sc. recalled) Characteristics:
\begin{enumerate}
\item \textbf{Emotional depth and intensity:} Entries in this category are characterized by a high level of emotional content, often describing feelings of sadness, joy, anxiety, or relief.

\item \textbf{Personal and introspective:} These entries often include personal reflections and introspection, exploring the emotional impact of events on the writer.

\item \textbf{Detailed and elaborate:} The descriptions are often more elaborate and detailed, providing a comprehensive view of the events and their emotional significance.

\item \textbf{Significant life events:} Entries frequently describe major life events, such as births, deaths, illnesses, or significant personal challenges.

\item \textbf{Narrative style:} The writing style is often more narrative, telling a story with a clear beginning, middle, and end, and focusing on the emotional journey.

\end{enumerate}

\subsection{Literary time}\label{gpt:literarytime}

[Note for this regression task we used the following prompt.]

\begin{figure}[h]
    \centering
\begin{tcolorbox}[colframe=black,colback=white,boxrule=0.5mm,arc=0mm]

Consider the data below, which contains a list of text/label pairs that illustrate a regression problem:

[10pt]
	[\{"text": "Carlyle disappeared. Barbara had taken off her own warm night-gown then, and put it upon William in place of his cold one--had struck a light and was busily dressing herself. \ldots", "label": -4.454347296253507\}
\\[10pt]

    \{"text": "At first he lived up above entirely, reading a good deal, and strumming upon an old harp \ldots", "label": 4.033719940405666\}

    \ldots ]
\\[10pt]
	Using this data, and this data alone, what are the textual characteristics that differentiate the two ends of this scale (let us call those ends ``Pole 1'' and the ``Pole 2'')?  The label ``-1'' is closer to Pole 1 than the label ``1'' is; the label ``1'', likewise, is closer to Pole 2 than the label ``-1'' is.  Provide a list of bullet points of those textual features in the following format:
\\[10pt]
	\#\#\# Pole 1 Characteristics:

	- **\\
	- **

	\ldots

	\#\#\# Pole 2 Characteristics:

	- **\\
	- **
\end{tcolorbox}
    \caption{Template for LLM sensemaking exercise for regression task.\label{regressionprompt}}
\end{figure}

\noindent Pole 1 (sc. shorter time) Characteristics:
\begin{enumerate}
\item \textbf{Frequent use of dialogue and conversational tone}: Texts often include direct speech, with characters interacting in a casual or informal manner.

\item \textbf{Presence of conflict or tension}: Many passages involve arguments, disagreements, or confrontations between characters.

\item \textbf{Focus on mundane or everyday events}: The content often revolves around ordinary activities, such as meals, conversations, or minor incidents.

\item \textbf{Detailed descriptions of characters' actions and emotions}: There is a strong emphasis on what characters are doing and feeling in the moment.

\item \textbf{Use of colloquial language and idiomatic expressions}: The language tends to be more relaxed and may include slang or regional dialects.

\item \textbf{Shorter, more fragmented sentences}: The writing style often includes brief, choppy sentences that reflect the immediacy of the scene.

\item \textbf{Frequent shifts in focus or perspective}: The narrative may jump between different characters or scenes, creating a sense of movement and dynamism.

\end{enumerate}
\noindent Pole 2 (sc. longer time) Characteristics:
\begin{enumerate}
\item \textbf{Formal and elaborate language}: The text often employs a more sophisticated and refined vocabulary, with complex sentence structures.

\item \textbf{Descriptive and expository passages}: There is a greater emphasis on setting the scene, providing background information, and describing the environment in detail.

\item \textbf{Philosophical or reflective tone}: The content may include introspective or contemplative passages, exploring deeper themes and ideas.

\item \textbf{Historical or literary references}: The text may allude to historical events, classical literature, or other cultural touchstones.

\item \textbf{Focus on significant or dramatic events}: The narrative often centers around major occurrences, such as battles, political intrigues, or life-changing decisions.

\item \textbf{Longer, more complex sentences}: The writing style tends to be more fluid and continuous, with fewer breaks and a more measured pace.

\item \textbf{Consistent narrative perspective}: The text usually maintains a steady point of view, providing a cohesive and unified narrative voice.

\end{enumerate}

\subsection{Narrativity}\label{gpt:narrativity}

\noindent Category 0 (sc. narrative) Characteristics:
\begin{enumerate}
\item \textbf{Narrative Style}: The texts often have a narrative or storytelling style, focusing on characters, events, and dialogues.

\item \textbf{Descriptive Language}: Rich in descriptive language, often detailing settings, emotions, and actions.

\item \textbf{Literary Elements}: Use of literary elements such as metaphors, similes, and personification.

\item \textbf{Character Interaction}: Frequent interactions between characters, including dialogues and internal monologues.

\item \textbf{Emotional Tone}: Often convey emotions and personal experiences, creating an immersive atmosphere.

\item \textbf{Historical or Fictional Context}: Many texts are set in historical or fictional contexts, providing a backdrop for the narrative.

\item \textbf{Plot Development}: Presence of a clear plot or storyline, with events unfolding over time.

\item \textbf{Imagery}: Use of vivid imagery to paint scenes and evoke sensory experiences.

\end{enumerate}
\noindent Category 1 (sc. non-narrative) Characteristics:
\begin{enumerate}
\item \textbf{Formal and Technical Language}: Use of formal, technical, and legal language, often related to business, contracts, or regulations.

\item \textbf{Structured Format}: Texts are often structured in a formal format, including sections, clauses, and bullet points.

\item \textbf{Objective Tone}: An objective, impersonal tone, focusing on facts, procedures, and instructions.

\item \textbf{Specific Terminology}: Use of specific terminology related to business, law, or technical fields.

\item \textbf{Lack of Narrative}: Absence of a narrative or storytelling style; instead, the focus is on conveying information or instructions.

\item \textbf{Contractual and Legal Content}: Frequent references to agreements, obligations, rights, and legal terms.

\item \textbf{Professional Context}: Texts are often set in a professional or corporate context, dealing with business operations, legal matters, or technical specifications.

\item \textbf{Precision and Clarity}: Emphasis on precision and clarity, ensuring that the information is unambiguous and easily understood.

\end{enumerate}

\subsection{Strangeness}

\noindent Category 0 (sc. strange) Characteristics:
\begin{enumerate}
\item \textbf{Technical or Scientific Language}: Texts often include technical or scientific terms, such as ``proton and neutron structure,'' ``viewer magnification,'' ``synthetic gravity generator,'' and ``molecular started biting at the fort’s ray screen.''

\item \textbf{Space and Technology Themes}: Many texts reference space, ships, and technology, e.g., ``space navigator,'' ``rocket ships,'' ``air ship,'' ``space speed,'' and ``space tug.''

\item \textbf{Descriptive of Actions and Objects}: Texts frequently describe actions or objects in a detailed manner, such as ``The ship plunged planetward,'' ``The boats slid into space,'' and ``The hull of the \_Enterprise\_ bulked huge in the guidance screen.''

\item \textbf{Formal and Objective Tone}: The tone is often formal and objective, focusing on factual descriptions and technical details, e.g., ``The machine replenishes the energy because it is set to maintain a certain energy level in the field.''

\item \textbf{Lack of Personal Interaction}: There is minimal personal interaction or dialogue between characters, with a focus on events and descriptions, e.g., ``The car roared through the gates of the spaceport and sped across the hard surface of the field.''

\end{enumerate}
\noindent Category 1 (sc. not strange) Characteristics:
\begin{enumerate}
\item \textbf{Personal and Emotional Language}: Texts often include personal and emotional language, such as ``She felt her chest tighten and her face become stiff,'' ``Her feeling of distress was much deeper,'' and ``He liked young people.''

\item \textbf{Dialogue and Interactions}: Many texts feature dialogue and interactions between characters, e.g., ``Oh you call it helping me do you he asked acidly,'' ``Good morning Alan she said,'' and ``What do you mean inquired Mr. Roumann.''

\item \textbf{Everyday Activities and Settings}: Texts frequently describe everyday activities and settings, such as ``She fished in the voluminous pockets of her jeans,'' ``Alan slurped down the rest of his coffee,'' and ``She rattled around in the galley for some time.''

\item \textbf{Informal and Conversational Tone}: The tone is often informal and conversational, focusing on personal experiences and interactions, e.g., ``I just know it’ll be fantastic,'' ``Well that’s awfully nice of you Nothing at all,'' and ``I m hungry.''

\item \textbf{Presence of Personal Pronouns}: There is a higher usage of personal pronouns, indicating a focus on individuals and their experiences, e.g., ``I thought you said your name was Alan Alan Adam Andy,'' ``I m sure we ll get along fine Thanks Natalie,'' and ``I m also getting a few things ready so we can have a fast breakfast in case we have to eat on the run.''

\end{enumerate}

\subsection{Stream-of-consciousness}

\noindent Category 0 (sc. realism) Characteristics:
\begin{enumerate}
\item \textbf{Dialogue-Driven}: Texts often contain significant amounts of dialogue between characters, which drives the narrative forward.

\item \textbf{Everyday Situations}: The scenarios described are often mundane or everyday occurrences, such as conversations about family, meals, or simple activities.

\item \textbf{Realistic Tone}: The tone tends to be straightforward and realistic, focusing on practical matters and direct interactions.

\item \textbf{Character Interactions}: Emphasis on interactions between characters, often involving social or familial relationships.

\item \textbf{Descriptive but Practical}: Descriptions are present but are more practical and less ornate, focusing on the immediate environment or actions.

\item \textbf{Conflict and Resolution}: Often includes a clear conflict and resolution within the passage, typically involving personal or social issues.

\item \textbf{Historical or Social Context}: Some texts provide a historical or social context, often reflecting on societal norms or personal histories.

\end{enumerate}
\noindent Category 1 (sc. soc) Characteristics:
\begin{enumerate}
\item \textbf{Introspective and Reflective}: Texts often delve into the inner thoughts and emotions of characters, providing a deep introspective look.

\item \textbf{Descriptive and Poetic}: Rich, ornate descriptions that create vivid imagery and evoke strong emotions.

\item \textbf{Philosophical and Abstract}: Themes can be more abstract, philosophical, or existential, exploring deeper meanings and human conditions.

\item \textbf{Stream-of-Consciousness}: Some texts use a stream of consciousness style, reflecting the continuous flow of thoughts and feelings.

\item \textbf{Symbolism and Metaphor}: Frequent use of symbolism and metaphor to convey complex ideas and emotions.

\item \textbf{Emotional Depth}: High emotional intensity, often focusing on personal struggles, desires, and inner conflicts.

\item \textbf{Literary and Artistic References}: References to literature, art, or historical events are more common, adding layers of meaning to the text.

\end{enumerate}

\section{Learning rates}\label{appendix:lr}

\begin{figure}[h]
    \centering
    \includegraphics[width=.7\linewidth]{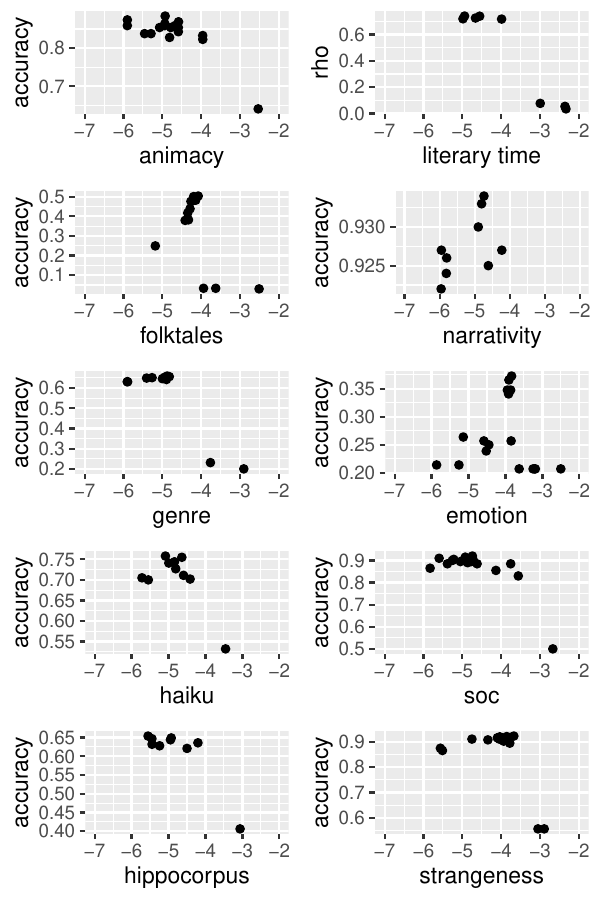}
    \caption{Learning rate sweep on development data for BERT across different tasks. The x axis is the $\log_{10}$ of the learning rate.}
    \label{bertlr}
\end{figure}

\begin{figure}[h]
    \centering
    \includegraphics[width=.7\linewidth]{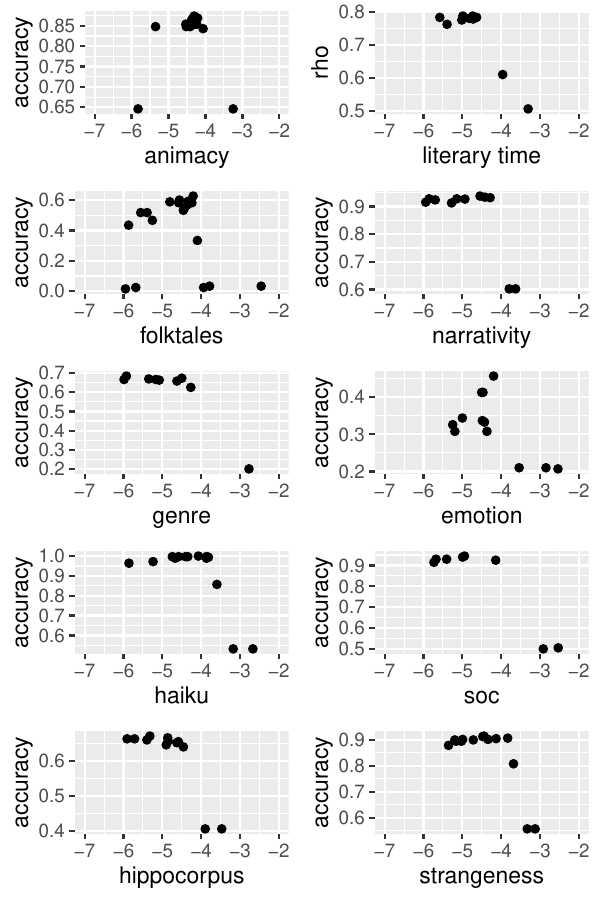}
    \caption{Learning rate sweep on development data for RoBERTa across different tasks. The x axis is the $\log_{10}$ of the learning rate.}
    \label{robertalr}
\end{figure}

\begin{figure}[h]
    \centering
    \includegraphics[width=.7\linewidth]{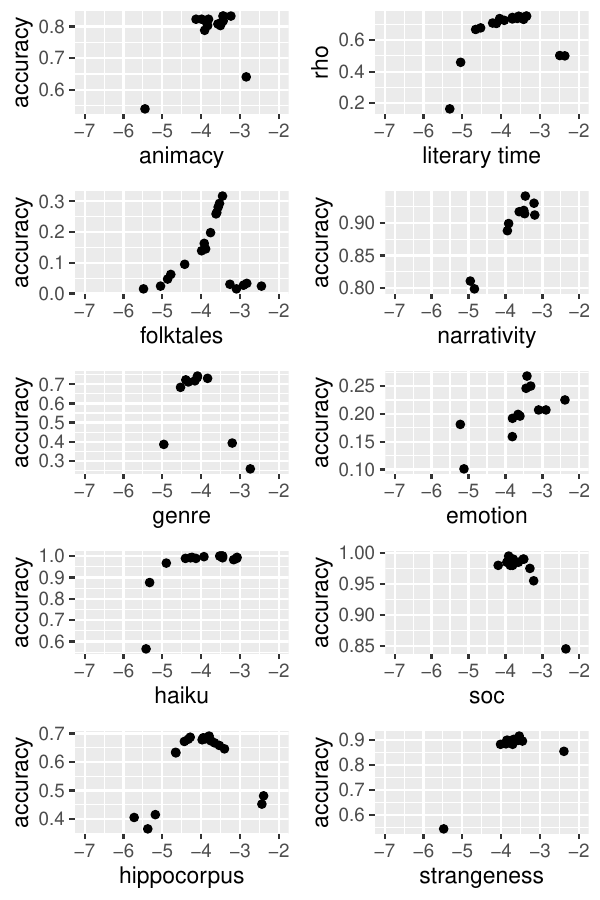}
    \caption{Learning rate sweep on development data for Llama 3 8B across different tasks. The x axis is the $\log_{10}$ of the learning rate.}
    \label{llamalr}
\end{figure}

Figures \ref{bertlr}, \ref{robertalr} and \ref{llamalr} illustrate the range of learning rates considered during hyperparameter optimization, along with the ensuing accuracy on development data; while we consider 50 optimization trials of each learning rate, \texttt{optuna} discards trials that are unlikely to outperform learning rates seen so far (so each task/model may have fewer than 50 points illustrated).  Across all supervised methods---BERT (fig. \ref{bertlr}), RoBERTa (fig. \ref{robertalr}) and Llama 8B (fig. \ref{llamalr})---the choice of learning rate is very dependent on the task, necessitating task-specific optimization.

\end{document}